\documentclass{article}

\usepackage{spconf}

\usepackage{graphicx}
\graphicspath{{fig/}}
\newcommand{\imagesep}{\vspace*{-6pt}}
\newcommand{\tablesep}{\vspace*{5pt}}

\usepackage{booktabs}
\usepackage{tabularx}
\usepackage{multirow}
\usepackage{siunitx}
\newcommand{\mytable}{
    \centering
    \renewcommand{\arraystretch}{1.1}
    }
\newcolumntype{C}{>{\centering\arraybackslash}X}
\newcolumntype{L}{>{\raggedright\arraybackslash}X}
\newcolumntype{R}{>{\raggedleft\arraybackslash}X}
\newcolumntype{P}[1]{>{\raggedright\arraybackslash}p{#1}}

\usepackage{amsmath}
\usepackage{amssymb}
\renewcommand{\vec}[1]{\boldsymbol{{#1}}}

\usepackage{microtype}
\usepackage{url}
\newcommand{\system}[1]{{\small \textsc{#1}}}
\newcommand\blfootnote[1]{\begingroup
                          \renewcommand\thefootnote{}\footnote{#1}
                          \addtocounter{footnote}{-1}
                          \endgroup}

\usepackage{cite}
\let\oldbibliography\thebibliography
\renewcommand{\thebibliography}[1]{\oldbibliography{#1}
                                   \setlength{\itemsep}{0.75pt}
                                   \vspace*{-0mm}}

\usepackage[prependcaption,textsize=scriptsize]{todonotes}
\definecolor{mycolor}{HTML}{FF6600}

\title{Multilingual acoustic word embedding models for \\ processing zero-resource languages}
\name{Herman Kamper$^1$ \qquad Yevgen Matusevych$^2$ \qquad Sharon Goldwater$^2$}
\address{$^1$E\&E Engineering, Stellenbosch University \& $^2$School of Informatics, University of Edinburgh \\
         {\small \tt kamperh@sun.ac.za, yevgen.matusevych@ed.ac.uk, sgwater@inf.ed.ac.uk}}

\begin{document}

\maketitle
\begin{abstract}
Acoustic word embeddings are fixed-dimensional representations of variable-length speech segments.
In settings where unlabelled speech is the only available resource, such embeddings can be used in ``zero-resource'' speech search, indexing and discovery systems.
Here we propose to train a single supervised embedding model on labelled data from multiple well-resourced languages and then apply it to unseen zero-resource languages.
For this transfer learning approach, we consider two multilingual recurrent neural network models: a discriminative classifier trained on the joint vocabularies of all training languages, and a correspondence autoencoder trained to reconstruct word pairs. 
We test these using a word discrimination task on six target zero-resource languages. When trained on seven well-resourced languages, both models perform similarly and outperform unsupervised models trained on the zero-resource languages.
With just a single training language, the second model works better, 
but performance depends more on the particular training--testing language pair.
\end{abstract}
\section{Introduction}

Current automatic speech recognition (ASR) systems use supervised models trained on large amounts of transcribed speech audio.
For many low-resource languages, however, it is difficult or impossible to collect such annotated resources.
Motivated by the observation that infants acquire language without hard supervision, studies into ``zero-resource'' speech technology have started to develop unsupervised systems that can learn directly from unlabelled speech audio~\cite{jansen+etal_icassp13,versteegh+etal_sltu16,dunbar+etal_asru17}.\blfootnote{Code: {\scriptsize \url{https://github.com/kamperh/globalphone_awe}}}

For zero-resource tasks such as query-by-example speech search, where the goal is to identify utterances containing a spoken query~\cite{hazen+etal_asru09,wang+etal_icassp18}, or full-coverage segmentation and clustering, where the aim is to tokenise an unlabelled speech set into word-like units~\cite{lee+etal_tacl15,elsner+shain_emnlp17,kamper+etal_asru17}, speech segments of different durations need to be compared.
Alignment methods such as dynamic time warping 
are computationally expensive and have limitations~\cite{rabiner+etal_tassp78}, so \textit{acoustic word embeddings} have been proposed as
an alternative:
a variable-duration speech segment is mapped to a fixed-dimensional vector so that instances of the same word type have similar embeddings~\cite{levin+etal_asru13}. Segments can then easily be compared by simply calculating a distance between their vectors in this embedding space.

Several supervised and unsupervised acoustic embedding methods have been proposed.
Supervised methods include convolutional~\cite{kamper+etal_icassp16,yuan+etal_interspeech18,haque+etal_icassp19} and recurrent neural network (RNN) models~\cite{settle+livescu_slt16,chung+glass_interspeech18,chen+etal_slt18,palaskar+etal_icassp19}, trained with discriminative classification and contrastive losses.
Unsupervised methods include using distances to a fixed reference set~\cite{levin+etal_asru13} and unsupervised autoencoding RNNs~\cite{chung+etal_interspeech16,audhkhasi+etal_stsp17,holzenberger+etal_interspeech18}.
The recent unsupervised RNN of~\cite{kamper_icassp19}, which we refer to as the correspondence autoencoder RNN (\system{CAE-RNN}), is trained on pairs of word-like segments found in an unsupervised way.
Unfortunately, while unsupervised methods are useful in that they can be used in zero-resource settings, there is still a large performance gap compared to supervised methods~\cite{kamper_icassp19}.
Here we investigate whether supervised modelling can still be used to obtain accurate embeddings on a language for which no labelled data is~available.

Specifically, we propose to exploit labelled resources from languages where these are available, allowing us to take advantage of supervised methods, but then apply the resulting model to zero-resource languages for which no labelled data is available.
We consider two multilingual acoustic word embedding models: a discriminative classifier RNN, trained on the joint vocabularies of several well-resourced languages, and a multilingual \system{CAE-RNN}, trained on true (instead of discovered) word pairs from the training languages.
We use seven languages from the GlobalPhone corpus~\cite{schultz+etal_icassp13} for training, and evaluate the resulting models on six {different} languages which are treated as zero-resource.
We show that supervised multilingual models consistently outperform unsupervised monolingual models trained 
on each of the target zero-resource languages.
When fewer training languages are used, the multilingual \system{CAE-RNN} generally performs better than the classifier, but performance is also affected by the 
{combination of training and test languages.} 

This study is inspired by recent work showing the benefit of using multilingual bottleneck features as frame-level representations for zero-resource languages~\cite{vesely2012language,yuan+etal_icassp17,hermann+etal_arxiv18,menon+etal_interspeech19}.
In~\cite{ondel+etal_arxiv19}, multilingual data were used in a similar way for discovering acoustic units.
As in those studies, our findings show the advantage of learning from 
labelled data in well-resourced languages when processing an unseen low-resource language---here at the word  rather than subword level.
Our work also takes inspiration from 
studies in multilingual ASR, where a single recogniser is trained to transcribe speech from any of several 
languages~\cite{tong+etal_arxiv17,toshniwal+etal_icassp18,cho+etal_slt18,adams+etal_arxiv19}. In our case, however, the language on which the model is applied is never seen during training;{ our approach is therefore a form of transfer learning~\cite{pan+yang_kde09}.}

\section{Acoustic word embedding models}

For obtaining acoustic word embeddings on a zero-resource language, we compare unsupervised models trained on within-language unlabelled data to supervised models trained on pooled labelled data from multiple well-resourced languages.
We use RNNs with gated recurrent units~\cite{chung+etal_arxiv14} throughout.

\subsection{Unsupervised monolingual acoustic embeddings} 
\label{sec:unsupervised_models}

\begin{figure}[!b]
    \centering
    \vspace*{-4pt}
    \includegraphics[scale=0.85]{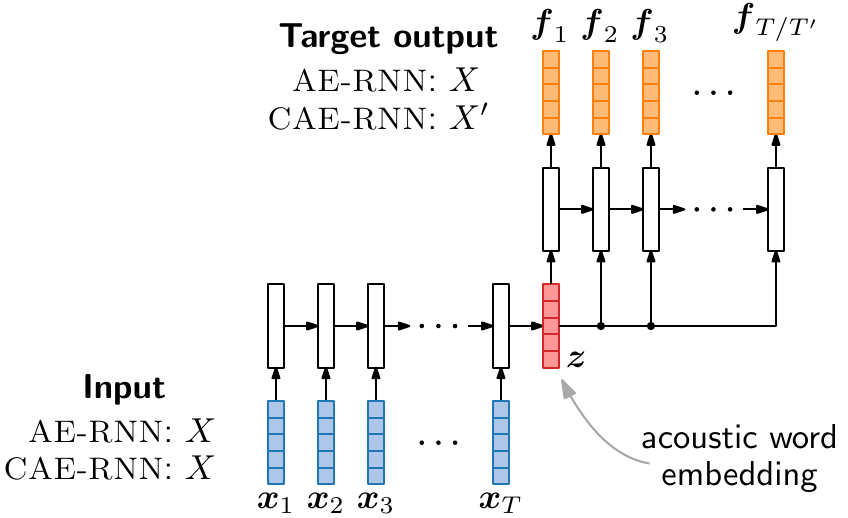}
    \imagesep
    \caption{The \system{AE-RNN} is trained to reconstruct its input $X$ (a speech segment) from the latent acoustic word embedding $\vec{z}$. The \system{CAE-RNN} uses an unsupervised term discovery system to find word pairs $(X, X')$, and is then trained to reconstruct one segment when presented with the other as input.}
    \label{fig:c_ae_rnn}
\end{figure}

We
consider two unsupervised monolingual embedding models. 
Both have an encoder-decoder RNN structure: an encoder RNN reads an input sequence and sequentially updates its internal hidden state, while a decoder RNN produces the output sequence conditioned on the final encoder output~\cite{sperduti+starita_tnn97,cho+etal_emnlp14}.

The unsupervised autoencoding RNN (\system{AE-RNN}) of~\cite{chung+etal_interspeech16} is trained on unlabelled speech segments to reproduce its input, as illustrated in Figure~\ref{fig:c_ae_rnn}.
The final fixed-dimensional output $\vec{z}$ from the encoder (red in Figure~\ref{fig:c_ae_rnn}) gives the acoustic word embedding.
Formally, an input segment $X = \vec{x}_1, \vec{x}_2, \ldots, \vec{x}_T$ consists of a sequence of frame-level acoustic feature vectors $\vec{x}_t \in \mathbb{R}^D$ (e.g.\ 
MFCCs).
The loss for a single training example is 
$\ell(X) = \sum_{t = 1}^T \left|\left|\vec{x}_t - \vec{f}_t(X)\right|\right|^2$,
with
$\vec{f}_t(X)$
the $t^\textrm{th}$ decoder output conditioned on the latent embedding $\vec{z}$. 
As in~\cite{audhkhasi+etal_stsp17}, we use a transformation of the final hidden state of the encoder RNN to produce the embedding $\vec{z} \in \mathbb{R}^M$. 

We next consider the unsupervised correspondence autoencoder RNN (\system{CAE-RNN}) of~\cite{kamper_icassp19}.
Since we do not have access 
to transcriptions, an
unsupervised term discovery~(UTD) system is applied to an unlabelled speech collection in the target zero-resource language, discovering pairs of word segments predicted to be of the same unknown type. 
These are then presented as input-output pairs to the \system{CAE-RNN}, as shown in Figure~\ref{fig:c_ae_rnn}.
Since UTD is itself unsupervised, the overall approach is unsupervised.
The idea is that the model's embeddings should be
invariant
to properties not common to the two segments (e.g.\ speaker,
channel), while capturing aspects that are (e.g.\ word identity).
Formally, a single training item consists of a pair of segments $(X, X')$.
Each segment consists of a unique sequence of frame-level vectors: {$X = \vec{x}_1, \ldots, \vec{x}_{T}$} and {$X' = \vec{x}'_1, \ldots, \vec{x}'_{T'}$}.
The loss for a single training pair is 
\mbox{$\ell(X, X') = \sum_{t = 1}^{T'} ||\vec{x}'_t - \vec{f}_t(X)||^2$},
where $X$ is the input and $X'$ the target output sequence.
We first pretrain the model using the AE loss above before switching to the CAE~loss.

\subsection{Supervised multilingual acoustic embeddings} 
\label{sec:supervised_models}

\begin{figure}[!b]
    \vspace*{-3pt}
    \centering
    \includegraphics[scale=0.85]{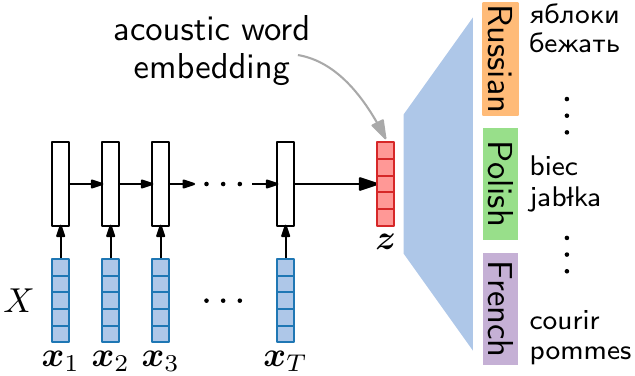}
    \imagesep
    \caption{The multilingual \system{ClassifierRNN} is trained jointly on all the training languages to classify which word type an input segment $X$ belongs to. Our model is trained on data from seven languages (three shown here for illustrative purposes).
    }
    \label{fig:classifier_rnn}
\end{figure}

Given labelled data from several well-resourced languages, we 
consider
supervised multilingual 
acoustic embedding models.

Instead of using discovered word segments in the target zero-resource language, multilingual \system{AE-RNN} and \system{CAE-RNN} models can be trained on the pooled ground truth word segments from forced alignments in the well-resourced languages.
Although these models are not explicitly discriminative, they do make use of ideal information 
and are therefore 
supervised.

As an alternative, we consider an explicitly discriminative model. 
Given a true word segment $X$ from any one of the training languages, the \system{ClassifierRNN} predicts the word type of that segment.
Formally, it is trained using the multiclass log loss, $\ell(X) = - \sum_{k = 1}^K y_k \, \log f_k(X)$, where $K$ is the size of the joint vocabulary over all the training languages, $y_k$ is an indicator for whether $X$ is an instance of word type $k$, and $\vec{f}(X) \in [0, 1]^K$ 
is the predicted distribution over the joint vocabulary.
An acoustic word embedding $\vec{z}$ is obtained from an intermediate layer shared between all training languages. This embedding is fed into a softmax layer to produce $\vec{f}(X)$, as illustrated in Figure~\ref{fig:classifier_rnn}.
Embeddings can therefore be obtained for speech segments from a language not seen during training.

We could also have used a contrastive loss, 
but the classifier performs only slightly worse in the supervised monolingual case~\cite{kamper+etal_icassp16,settle+livescu_slt16}. It is also 
easier to extend 
as it does not require a procedure for sampling pairs over multiple languages.

\section{Experimental setup}

We perform 
our experiments on the GlobalPhone corpus of read speech~\cite{schultz+etal_icassp13}.
As in~\cite{hermann+etal_arxiv18}, we treat six languages as 
our target
zero-resource
languages: 
Spanish (ES), Hausa (HA), Croatian (HR), Swedish (SV), Turkish (TR) and Mandarin (ZH).
Each language has on average 16 hours of training, 2 hours of development and 2 hours of test data.
Since we do not {use}
transcriptions for the unsupervised monolingual embedding models (\S\ref{sec:unsupervised_models}), we apply the UTD system of~\cite{jansen+vandurme_asru11} to each of the training sets, and use the discovered word segments to train unsupervised monolingual \system{CAE-RNN} models on each language.\footnote{For extracting pairs, we specifically use the code and steps described at:  {\scriptsize \url{https://github.com/eginhard/cae-utd-utils}}.}
Roughly 36k predicted word pairs are extracted in each language.
\system{AE-RNN} models are similarly trained on the UTD-discovered segments (this gives slightly better performance than training on random segments\cite{kamper_icassp19}).

For training supervised multilingual embedding models~(\S\ref{sec:supervised_models}), seven
other GlobalPhone languages are chosen as well-resourced languages: 
Bulgarian (BG), Czech (CS), French (FR), Polish (PL), Portuguese~(PT), Russian (RU) and Thai (TH).
Each language has on average 21 hours of labelled training data.
We pool the data from all languages and train supervised multilingual variants of the \system{AE-RNN} and \system{CAE-RNN} using true word segments and true word pairs 
obtained from forced alignments.
Rather than considering all possible word pairs from all languages when training the multilingual \system{CAE-RNN}, we sample 300k true word pairs from the combined data. Using more pairs did not improve development performance, but increased training time. 
The multilingual \system{ClassifierRNN} is trained jointly on true word segments from the seven training languages.
The number of word types per language is limited to 10k, giving a total of 70k output classes (more classes did not give improvements).

Since development data would not 
be available in a zero-resource language, 
we performed development experiments on labelled
data from yet another language: German. 
We used this 
data to tune the number of pairs for the \system{CAE-RNN}, the vocabulary size for the \system{ClassifierRNN} and the number of training epochs.
Other hyperparameters are set as in~\cite{kamper_icassp19}.

All models are implemented in TensorFlow. 
Speech audio is parametrised as $D = \text{13}$ dimensional static Mel-frequency cepstral coefficients (MFCCs).
We use an embedding dimensionality of $M = \text{130}$ throughout, since downstream systems such as the segmentation and clustering system of~\cite{kamper+etal_asru17} are
constrained to embedding sizes of this order.
All encoder-decoder models have 3 encoder and 3 decoder
unidirectional RNN layers, each with 400 units.
The same encoder structure is used for the \system{ClassifierRNN}.
Pairs are presented to the \system{CAE-RNN} models in both input-output directions.
Models are trained using
Adam optimisation 
with a learning rate of 0.001.

We want to measure intrinsic acoustic word embedding quality without being tied to a particular downstream
system architecture. We therefore use a word discrimination
task designed for this purpose~\cite{carlin+etal_icassp11}. In the \textit{same-different} task,
we are given a pair of acoustic segments, each a true word,
and we must decide whether the segments are examples of
the same or different words. To do this using an embedding
method, a set of words in the test data  are embedded using a particular 
approach. For every word pair in this set, the cosine distance
between their embeddings is calculated. Two words can then
be classified as being of the same or different type based on
some threshold, and a precision-recall curve is obtained by
varying the threshold. The area under this curve is used as final
evaluation metric, referred to as the average precision (AP).
We use the same specific evaluation setup as in~\cite{hermann+etal_arxiv18}.

As an additional unsupervised baseline embedding method, we use downsampling~\cite{holzenberger+etal_interspeech18}
by keeping 10 equally-spaced MFCC vectors from a segment
with appropriate interpolation, giving a 130-dimensional embedding.
Finally, we report same-different performance when using
dynamic time warping (DTW) alignment cost
as a score for word discrimination.

\section{Experimental results}

Table~\ref{tbl:results_test} shows the performance 
of the unsupervised monolingual and supervised multilingual models applied to
test data from the six zero-resource languages.
Comparing the unsupervised techniques, the \system{CAE-RNN} outperforms downsampling and the \system{AE-RNN} on all six languages, as also shown in~\cite{kamper_icassp19} on English and Xitsonga data.
It even outperforms DTW on Spanish, Croatian and Swedish; this is noteworthy since DTW uses full alignment 
to discriminate between words (i.e.\ it has access to the full sequences without any compression).

Comparing the best unsupervised model (\system{CAE-RNN}) to the supervised multilingual models (last two rows), we see that the 
multilingual models consistently perform better across all six zero-resource languages.
The relative performance of the supervised multilingual \system{CAE-RNN} and \system{ClassifierRNN} models is not consistent over the six zero-resource evaluation languages, with one model working better on some languages while another works better on others.
However, both consistently outperform the unsupervised monolingual models trained directly on the target languages, showing the benefit of incorporating data from languages where labels are available.

\begin{table}[!t]
    \mytable
    \caption{AP (\%) on test data for the zero-resource languages. The unsupervised \system{CAE-RNN}s are trained separately for each zero-resource language on segments from a UTD system applied to unlabelled monolingual data.
    The multilingual models are trained on ground truth word segments obtained by pooling labelled training data from seven well-resourced languages.}
    \tablesep
    \begingroup
    \small
    \begin{tabularx}{1.0\linewidth}{@{}L c S[table-format=2.1] cc S[table-format=2.1] c@{\ }}
        \toprule
        Model & ES & HA & HR & SV & {TR} & ZH \\
        \midrule
        \multicolumn{4}{@{}l}{\underline{\textit{Unsupervised {models}:}}} \\[2pt]
        \system{DTW} & 29.7 & 20.1 & 13.7 & 24.2 & 11.9 & 27.1 \\
        \system{Downsample} & 19.4 & 10.7 & 11.2 & 16.6 & 8.0 & 20.2 \\
        \system{AE-RNN} (UTD) & 18.1 & 6.5 & 10.4 & 12.0 & 6.8 & 18.5 \\
        \system{CAE-RNN} (UTD) & 39.7 & 17.8 & 21.4 & 25.2 & 10.7 & 21.3 \\[2pt]
        \multicolumn{4}{@{}l}{\underline{\textit{Multilingual {models}:}}} \\[2pt]
        \system{CAE-RNN} & \textbf{56.0} & \textbf{32.7} & 29.9 & \textbf{36.7} & 20.9 & 34.2 \\
        \system{ClassifierRNN} & 54.3 & 29.5 & \textbf{32.9} & 33.5 & \textbf{21.2} & \textbf{34.5} \\
        \bottomrule
    \end{tabularx}
    \endgroup
    \label{tbl:results_test}
\end{table}

\begin{figure}[!t]
    \centering
    \includegraphics[width=0.99\linewidth]{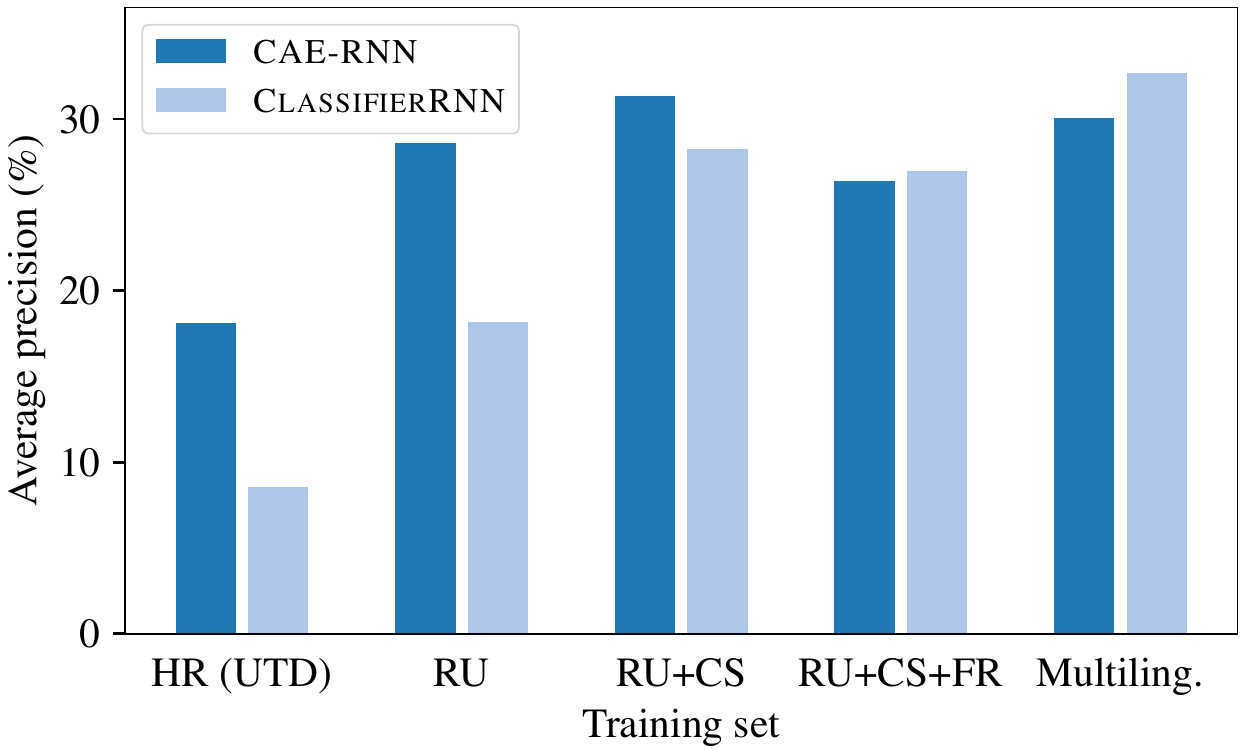}
    \imagesep
    \vspace*{-4pt}
    \caption{AP on Croatian (HR) development data for \system{CAE-RNN} and \system{ClassifierRNN} models as more training languages are added. The `multiling.'\ models are trained on all seven well-resourced languages. 
    Scores when training on UTD segments (extracted from unlabelled HR data) are given as a reference. 
    }
    \label{fig:cr_bar}
    \vspace*{-3pt}
\end{figure}

We are also interested in determining the effect of using fewer training languages.
Figure~\ref{fig:cr_bar} shows development performance on a single evaluation language, Croatian, as supervised models are trained 
on one, three and all seven well-resourced languages.
We found similar patterns with
all six zero-resource languages, but only show the Croatian results here.
In general the \system{CAE-RNN} outperforms the \system{ClassifierRNN} when fewer training languages are used.
Using labelled data from a single training language (Russian) already outperforms unsupervised monolingual models trained on Croatian UTD segments.
Adding an additional language (Czech) further improves AP.
However, when adding French (RU+CS+FR), performance drops; this was not the case on most of the other zero-resource languages, but is specific to Croatian.
Croatian is in the same language family as Russian and Czech, {which may explain this effect}.
When all languages are used (right-most bar), the multilingual \system{CAE-RNN} performs similarly to the RU+CS case, while the \system{ClassifierRNN} performs slightly better.

\begin{figure}[!t]
    \centering
    \includegraphics[width=0.525\linewidth]{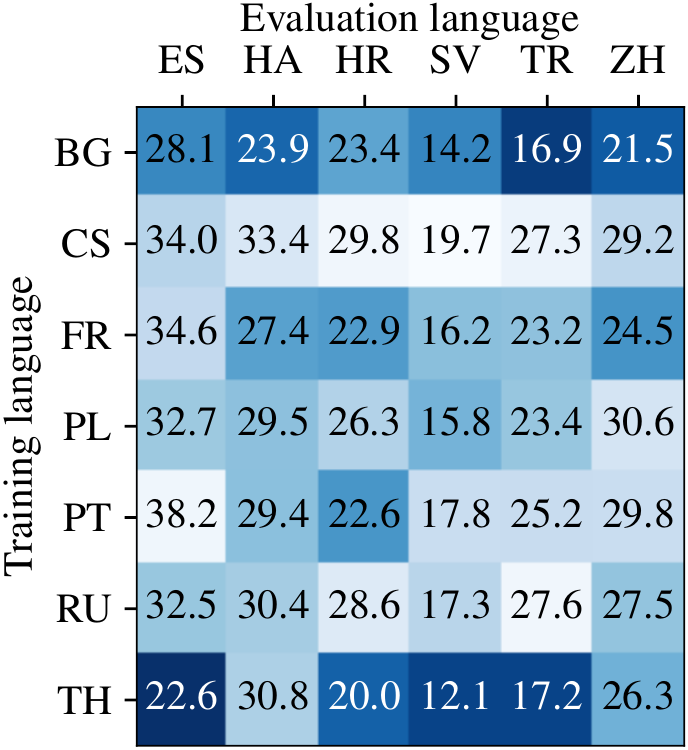}
    \imagesep
    \caption{AP (\%) on development data for the six zero-resource languages (columns) when applying different monolingual \system{CAE-RNN} models, each trained on labelled data from a well-resourced language (rows). Heatmap colours are normalised for each zero-resource language (i.e.\ per column).
    }
    \vspace*{-3pt}
    \label{fig:crosslingual}
\end{figure}

To further investigate the impact of training language choice, we train supervised monolingual \system{CAE-RNN} models on each of the training languages, and then apply each model to each of the zero-resource languages.
Results are shown in Figure~\ref{fig:crosslingual}.
We observe that the choice of well-resourced language can greatly impact performance.
On Spanish, using Portuguese is better than any other language, and similarly on Croatian, the monolingual Russian and Czech systems perform well, showing that training on languages from the same family is beneficial.
Although performance can differ dramatically based on the source-target language pair, all of the systems in Figure~\ref{fig:crosslingual} outperform the unsupervised monolingual \system{CAE-RNN}s trained on UTD pairs from the target language (Table~\ref{tbl:results_test}), except for the BG-HA pair.
Thus, training on just a single well-resourced language is beneficial in almost all cases (a very recent study \cite{yang+hirschberg_interspeech19} made a similar finding).
Furthermore, the multilingual \system{CAE-RNN} trained on all seven languages (Table~\ref{tbl:results_test}) outperforms the supervised monolingual \system{CAE-RNN}s for all languages, apart from some systems applied to Turkish.
The performance effects of language choice therefore diminish as more training languages are used.

\section{Conclusion}

We proposed to train supervised multilingual acoustic word embedding models by pooling labelled data from well-resourced languages.
We applied models trained on seven languages to six zero-resource languages without any labelled data.
Two multilingual models (a discriminative classifier and a correspondence model 
with a reconstruction-like loss) consistently 
outperformed monolingual unsupervised model trained directly on the 
zero-resource language.
When fewer training languages are used, we showed that the correspondence recurrent neural network outperforms the classifier and 
that performance is affected by the combination of training and test languages.
These effects diminish as more training languages are used. 
Future work will analyse the types of language-independent properties
captured through this transfer learning~approach.\vspace*{-3pt}\blfootnote{
This work is based on research supported in part by
the National Research Foundation of South Africa (grant number:\ 120409), a James S.\ McDonnell Foundation Scholar Award (220020374), an ESRC-SBE award (ES/R006660/1), and a Google Faculty Award for HK.}


\newpage
\ninept
\bibliography{mybib_short}

\end{document}